\documentclass[runningheads]{llncs}
\usepackage[T1]{fontenc}
\usepackage{adjustbox}
\usepackage{algorithm}
\usepackage{algpseudocode}
\usepackage{amsmath}
\usepackage{amssymb}
\usepackage{bbm}
\usepackage{booktabs}
\usepackage{colortbl}
\usepackage{epsfig}
\usepackage{enumerate}
\usepackage{filecontents}
\usepackage{framed}
\usepackage{graphicx}
\usepackage[hidelinks]{hyperref}
\usepackage{multirow}
\usepackage{orcidlink}
\usepackage{pgfplotstable}
\usepackage{pgfplots}
\usepackage{subcaption}
\usepackage{tabularx}
\usepackage{tikz}
\usepackage{tikzscale}
\usepackage{times}

\usepackage[mode=buildnew]{standalone} % requires -shell-escape
\usetikzlibrary{arrows}
\usetikzlibrary{calc}
\usetikzlibrary{positioning, fit, shapes.geometric, backgrounds}

\newcolumntype{Y}{>{\centering\arraybackslash}X}

\pgfplotsset{compat=1.7}

\graphicspath{{images/}}

\makeatletter
\renewcommand{\fnum@figure}{Fig. \thefigure}
\makeatother

\makeatletter
\renewcommand{\fnum@table}{Table \thetable}
\makeatother

\definecolor{JUNGLE_GREEN}{RGB}{0, 178, 149}
\definecolor{CAROLINA_BLUE}{RGB}{27, 152, 224}
\hypersetup{
  colorlinks=true,
  citecolor=CAROLINA_BLUE
}

\begin{document}

\title{Person Re-Identification via\\ Generalized Class Prototypes}

\titlerunning{Generalized Class Prototypes}

\author{Md Ahmed Al Muzaddid\orcidlink{0000-0002-7916-714X} \and William J.
Beksi\orcidlink{0000-0001-5377-2627}}

\institute{The University of Texas at Arlington, Arlington TX 76019, USA}

\maketitle              

%%%%%%%%%%%%%%%%%%%%%%%%%%%%%%%%%%%%%%%%%%%%%%%%%%%%%%%%%%%%%%%%%%%%%%%%%%%%%%%%
\begin{abstract}
Advanced feature extraction methods have significantly contributed to enhancing
the task of person re-identification. In addition, modifications to objective
functions have been developed to further improve performance. Nonetheless,
selecting better class representatives is an underexplored area of research
that can also lead to advancements in re-identification performance.  Although
past works have experimented with using the centroid of a gallery image class
during training, only a few have investigated alternative representations
during the retrieval stage. In this paper, we demonstrate that these prior
techniques yield suboptimal results in terms of re-identification metrics. To
address the re-identification problem, we propose a generalized selection
method that involves choosing representations that are not limited to class
centroids. Our approach strikes a balance between accuracy and mean average
precision, leading to improvements beyond the state of the art. For example,
the actual number of representations per class can be adjusted to meet specific
application requirements. We apply our methodology on top of multiple
re-identification embeddings, and in all cases it substantially improves upon
contemporary results.


\end{abstract}

%%%%%%%%%%%%%%%%%%%%%%%%%%%%%%%%%%%%%%%%%%%%%%%%%%%%%%%%%%%%%%%%%%%%%%%%%%%%%%%%
\section{Introduction}
\label{sec:introduction}
\begin{figure}
\centering
\includegraphics[scale=1.0]{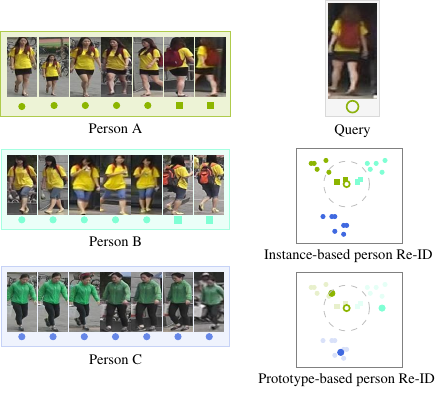}
\caption{An overview of prototype-based Re-ID. The solid circles/squares denote
the feature vectors of each image instance, while the empty circles represent
query vectors. The two plots on the right depict the feature vector distribution
in the embedding space. The upper-right plot illustrates instance-based Re-ID,
which typically yields lower precision. The plot depicts a specific case where
among the five nearest instances from the query, two are false positives. The
lower-right plot demonstrates how our prototype-based Re-ID approach enhances
precision by representing all instances from the same class using a
distribution-aware feature vector.}
\label{fig:overview}
\vspace{-1mm}
\end{figure}

Person re-identification (Re-ID) has been widely studied as a specific image
retrieval problem. Given a query image of a person, the goal is to identify it
from a set of gallery images captured by a group of non-overlapping cameras.
The gallery images typically include disjoint views of the same person taken by
different cameras at distinct times. Re-ID research has undergone rapid growth
since the introduction of initial datasets
\cite{ess2007depth,gray2007evaluating} for this task. The technology has found
widespread application in various domains such as autonomous vehicles, security
and surveillance systems, sports analytics, and much more. The Re-ID task is
challenging due to a combination of dynamic lighting conditions, low image
resolution, multiple camera viewpoints, occlusions, unconstrained poses, and
unreliable bounding boxes.

A common approach to person Re-ID is to first transform the query and gallery
images into feature vectors using either handcrafted feature engineering or
deep learning feature extractors. Subsequently, the similarity between the
query and gallery feature vectors is assessed by measuring the distance between
them. The greater the proximity of these vectors, the higher the degree of
similarity.  More specifically, we expect that the feature vectors representing
the same person will exhibit a close spatial relationship, while vectors
representing different individuals will have a significant separation in the
vector space where distance is measured by the Euclidean norm.

Since images for person Re-ID are typically transformed into feature vectors,
feature extraction via deep learning has become a key aspect in recent
advancements.  This research has focused on developing more discriminative
features, yet it has inherent limitations. Specifically, it is restricted to
extracting viewpoint-centric features from a single image, which restricts its
ability to form a comprehensive, person-centric representation by utilizing
multiple images of the same individual. Furthermore, the extensive image
comparisons required for Re-ID impose high computational demands, restricting
its practical application in resource-constrained environments.

As an alternative approach, Snell et al. \cite{snell2017prototypical} introduced
the prototypical network for Re-ID, a method that learns a metric space in which
classification is performed by calculating distances to prototype
representations for each class. Building on this paradigm, we propose a
generalization of this concept by employing multiple prototypes per class. We
show that using multiple prototypes per class enhances the performance of Re-ID
tasks compared to a single prototype per class. \textit{To the best of our
knowledge, we are the first to establish the effectiveness of varying numbers of
prototypes for image retrieval}.

Not only do we examine prototypes as class representatives, but we also
generalize the class representative selection process. In particular, we develop
a transformer decoder-based model that takes a set of images of an object of
interest and generates one or more set/class representatives in the embedding
space during inference. When generating multiple prototypes iteratively, the
decoder's self-attention module attends to all prototypes generated in preceding
iterations, while the cross-attention module is conditioned on the overall
feature distributions of the entire class.  Fig.~\ref{fig:overview} illustrates
the selection of class representatives via our method compared to conventional
(e.g., instance-based) approaches.

To demonstrate the effectiveness of our approach, we present baseline algorithms
for efficiently identifying robust prototypes. These methods sample class
representatives that capture intra-class diversity while maintaining inter-class
boundaries within the embedding space. In summary, our contributions are as
follows.
\begin{itemize}
  \item We provide an analysis of the impact of different class representatives
  (e.g., instance, centroid) on person Re-ID tasks.
  \item We create an attention-based model to generate class prototypes that
  can be directly compared with the query during inference time.
  \item We present multiple sampling-based algorithms for selecting class
  representatives and establish state-of-the-art person Re-ID benchmarks.
\end{itemize}
Our source code is publicly available at \cite{gcp}.

%%%%%%%%%%%%%%%%%%%%%%%%%%%%%%%%%%%%%%%%%%%%%%%%%%%%%%%%%%%%%%%%%%%%%%%%%%%%%%%%
\vspace{-1mm}
\section{Related Work}
\label{sec:related_work}
%-------------------------------------------------------------------------------
\subsection{Prototype-Based Methods}
Similar to the prototypical network approach introduced by Snell et al.
\cite{snell2017prototypical}, Wang et al. \cite{wang2019centroid} developed
embedding models that classify each query example by calculating distances to
speaker prototypes represented by centroids. Other research has explored
alternative strategies for aggregating image features. For example, centroid
vectors have been utilized during model training across various applications
\cite{lagunes2020centroids,yuan2020defense,zhang2021beyond}.

Recently, a centroid-based approach was used to represent gallery classes
during inference for person Re-ID \cite{wieczorek2021unreasonable}. This
improves performance by summarizing all gallery images of a single class into
an average centroid vector, enabling similarity measurement through distances
between query features and representations. However, we do not rely on a single
fixed prototype such as a centroid. Instead, our approach allows flexibility in
selecting a predefined number of representative points within the embedding
space based on the feature distribution.

%-------------------------------------------------------------------------------
\subsection{Attention Models}
Attention models have become very effective for person Re-ID. They can handle the part
alignment challenge and enhance feature representation. For instance, Liu et al.
\cite{liu2017end} proposed an end-to-end comparative attention network that
learns to focus on part pairs of images of people to compare their appearance.
Researchers have also employed attention models over a sequence of frames to
extract salient features \cite{liu2017quality,li2018diversity,si2018dual}. More
recently, Zhang et al. \cite{zhang2023pha} developed a patch-wise augmentation
technique to enhance the representation of high-frequency components in
attention-based models.

In this work, given a set of samples with their corresponding camera ID, we use
an attention model to find prototypes that represent a class. Our approach
exhibits parallels with prior works such as
\cite{zhong2017re,he2021transreid,wang2022nformer}, in which the inference time
setting allows the model to incorporate group information. Nevertheless, these
past works modify individual instances based on the group information while we
utilize this information to provide a reduced number of new prototypes.

%-------------------------------------------------------------------------------
\subsection{Loss Functions}
A substantial body of research delves into various loss functions for Re-ID.
The center loss, developed by Wen et al. \cite{wen2016discriminative},
facilitates concurrent learning of feature centers for each class and
effectively constrains large distances between features and their respective
class centers. A quadruplet loss function able to model an output with a larger
inter-class variation and a smaller intra-class variation, when compared to the
triplet loss, was presented by Chen et al. \cite{chen2017beyond}. Additionally,
there is the pair-wise contrastive \cite{chung2017two} and triplet ranking
\cite{zhong2018camera} losses. Zhu et al. \cite{zhu2020hetero} focused on the
heterogeneity of the data and developed a hetero-center loss to reduce
intra-class cross-modality variations. In contrast to dense comparisons that
use only a select number of suitable pairs for each class within a mini-batch,
a sparse pair-wise loss method was proposed by Zhou et al.
\cite{zhou2023adaptive}. We implement a custom loss function derived from the
triplet and contrastive losses to enforce larger inter-class and smaller
intra-class distances among the prototypes.

%%%%%%%%%%%%%%%%%%%%%%%%%%%%%%%%%%%%%%%%%%%%%%%%%%%%%%%%%%%%%%%%%%%%%%%%%%%%%%%%
\vspace{-1mm}
\section{Method}
\vspace{-4mm}
\label{sec:method}

\begin{figure}
\centering
\begin{subfigure}{.30\textwidth}
  \centering
  %\includestandalone[width=.97\linewidth, height=3cm]{images/instance}
  \includegraphics[width=.97\linewidth, height=3cm]{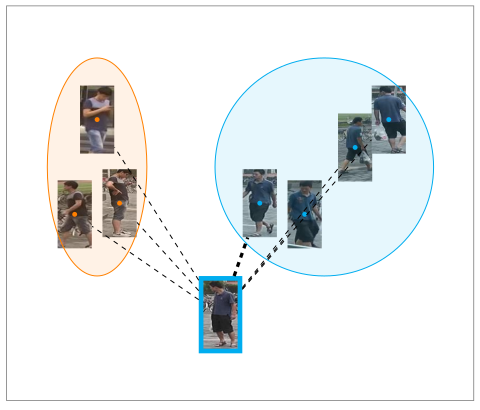}
  \caption{}
  \label{fig:instance_based}
\end{subfigure}
\hspace{10pt}
\begin{subfigure}{0.30\textwidth}
  \centering
  %\includestandalone[width=.97\linewidth, height=3cm]{images/centroid}
  \includegraphics[width=.97\linewidth, height=3cm]{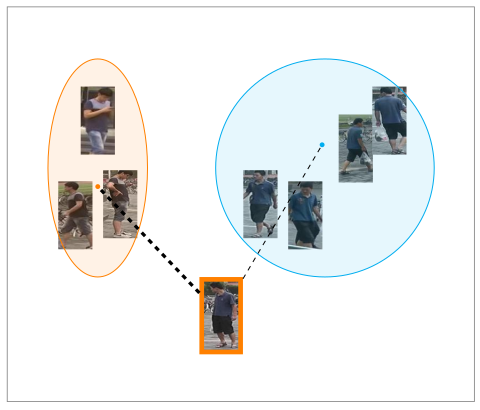}
  \caption{}
  \label{fig:centroid_based}
\end{subfigure}
\hspace{10pt}
\begin{subfigure}{0.30\textwidth}
  \centering
  %\includestandalone[width=.97\linewidth, height=3cm]{images/prototype}
  \includegraphics[width=.97\linewidth, height=3cm]{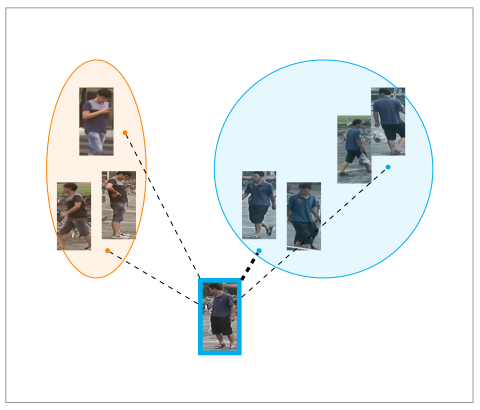}
  \caption{}
  \label{fig:prototype_based}
\end{subfigure}
\caption{A comparison among (a) instance-based, (b) centroid-based, and (c)
prototype-based Re-ID. The images enclosed by the colored rectangles are the
query images and the small colored dots are the class representatives. The
dashed lines indicate the distance between the query image and the class
representation. In (a), (b), and (c), the query image is assigned to the class
of the nearest instance, centroid, and prototype (our method) respectively.}
\label{fig:re-id_comparison}
\vspace{-8mm}
\end{figure}

\subsection{Motivation}
We first motivate our work by analyzing the decision boundaries and hypothesis
spaces associated with instance-based and centroid-based Re-ID methods. We
begin with the instance-based method, a case where each gallery image is
represented by its feature vector (Fig.~\ref{fig:instance_based}). In this
approach, the class label for a query image is assigned based on the most
similar gallery image, with similarity measured by calculating the distance
between the query image's feature vector and those of all the gallery images.
For the instance-based method, the decision boundary can be described in terms
of the gallery instances,
\begin{equation}
  \left\Vert x^{c_i}-b\right\Vert = \left\Vert x^{c_j} -b\right\Vert,
\end{equation}
where $b$ is a point on the decision boundary, and $x^{c_i}$  and $x^{c_j}$ are
the two closest gallery instances from $b$, representing two distinct classes
(i.e.,  $c_i$ and $c_j$). This decision boundary can become quite complex
depending on the distribution of the gallery instances.

The associated hypothesis space has a Vapnik-Chervonenkis (VC) dimension
\cite{vapnik2013nature} of $\mathcal{O}(n)$, where $n = |G|$ is the cardinality
of the gallery set. In Re-ID tasks, where $n \gg 100$, the VC dimension of the
instance-based approach increases substantially with the size of the gallery
set. This expansion of the hypothesis space makes the decision boundary more
susceptible to overfitting, which often results in a reduction of mean
precision.

Conversely, the centroid-based Re-ID approach (Fig.~\ref{fig:centroid_based})
imposes a highly-constrained hypothesis space that is restricted to only linear
class boundaries of the form $w\cdot x + b = 0$, where $x$ represents the
centroid, and $w$ and $b$ are the coefficients defining the separating
hyperplane. Due to the limited expressive capacity of this hypothesis space, it
often underfits the gallery instances. Thus, an intermediate approach that
strikes a balance between the overfitting tendencies of instance-based methods
and the underfitting limitations of centroid-based techniques is needed.

%-------------------------------------------------------------------------------
\subsection{Generalized Class Prototype}
Image similarity is typically measured by calculating the distance between a
query and a class representative (e.g., centroid) within a feature space.  We
refer to the class representative as a \textit{class prototype} and define the
retrieval process based on these prototypes as \textit{prototype-based image
retrieval}. Concretely, a class prototype can be any vector within the feature
space for which there exists a function,
\begin{equation}
  \mathcal{D}(p, q) = d \in \mathbb{R},
\end{equation}
that measures the distance $d$ between the query $q\in\mathbb{R}^n$ and the
prototype $p\in\mathbb{R}^n$. The number of prototypes, $N$, can exceed one
per class and may vary across different classes. We use the terms
\textit{class}, \textit{individual}, and \textit{identity} interchangeably.

Using these concepts, we define \textit{generalized class prototype (GCP)}
Re-ID as an image retrieval method in which a query image is re-identified by
measuring its similarity to class prototypes (Fig.~\ref{fig:prototype_based}).
The centroid, which is the mean of all the gallery images of a class, is a
special prototype. The Re-ID approach that quantifies similarity by measuring
the distance between the query and the centroid is a specific instance of
prototype-based image retrieval, where the number of prototypes per class is
fixed (i.e., $N = 1$). We refer to this as the \textit{centroid prototype}.

Traditional instance-based comparison, where each gallery image is individually
compared with a query image, is also a special case of prototype-based
retrieval but with a variable number of prototypes. More formally, the number
of prototypes per class is denoted as $N^{c_1}, N^{c_2}, \ldots, N^{c_m}$ where
$N^{c_1} = |G_{c_1}|, N^{c_2} = |G_{c_2}|, \ldots, N^{c_m} = |G_{c_m}|$.
$G_{c_i}$ represents the set of gallery images belonging to class $c_i$ and $m$
is the total number classes in the gallery set. We refer to this type of
prototype as the \textit{instance prototype}.

Given the special cases of prototype-based Re-ID, we can now address the
solution to the Re-ID problem in terms of a GCP. The objective is to assign
prototypes $p_k^{c_i}$ for any given query $q^{c_i}$ associated with the true
identity $c_i$ such that the following condition holds,
\begin{equation}
  \exists_{k \in \{1,\ldots,N^{c_i}\}}, \forall_{c_j: c_j\neq c_i}  \left\Vert p^{c_i}_k - q^{c_i} \right\Vert < \left\Vert p^{c{_j}}_{k^\prime} - q^{c_i}\right\Vert,
  \label{eqn:prototype_distance_query}
\end{equation}
where $p^{c_j}_{k^\prime}$ represents any prototype for identity $c_j$.
Without an exact query in advance, we assume that the gallery and query images
are independent and identically distributed. Under this assumption, gallery
images can serve as proxies for the query images. This allows us to substitute
$q^{c_i}$ with $x^{c_i}$ in \eqref{eqn:prototype_distance_query} resulting in
\begin{equation}
  \exists_{k \in \{1,\ldots,N^{c_i}\}}, \forall_{c_j: c_j\neq c_i}  \left\Vert p^{c_i}_k - x^{c_i}\right\Vert < \left\Vert p^{c_j}_{k^\prime} - x^{c_i} \right\Vert.
   \label{eqn:prototype_distance_gallery}
\end{equation}

A trivial solution to this inequality emerges if we assign $p^{c_i}_k =
x^{c_i}$. This is similar to the instance-based Re-ID scenario where the number
of prototypes per class is set to $N^{c_1} = |G_{c_1}|, N^{c_2} = |G_{c_2}|,
\ldots, N^{c_m} = |G_{c_m}|$. Restricting the number of prototypes per class to
$N^{c_i} < |G_{c_i}|$ makes solving \eqref{eqn:prototype_distance_gallery} a
non-trivial task. With a general gallery distribution and for any $N^{c_i} <
|G_{c_i}|$, an exact solution may not always be achievable.

To solve this problem, we optimize the prototype selection process. Rather than
seeking an exact solution, our goal is to minimize Re-ID errors. We employ an
attention-based model that effectively captures the structure of the gallery
distribution, enabling the generation of a set of prototypes per class. The
number of prototypes per class is treated as a tunable hyperparameter. To
reduce the number of hyperparameters, we constrain the model to generate a
fixed number of prototypes for all classes, i.e., $\forall_i N^{c_i} = N$.  For
any class where $|G_{c_i}| < N$, the number of prototypes can be adjusted by
setting $N^{c_i} =min(N, |G_{c_i}|)$.
%To mitigate overfitting, we further restrict the number of prototypes per
%class to $N \ll |G_c|$.

%-------------------------------------------------------------------------------
\subsection{Attention-Based Model}
\begin{figure}
\centering
\includegraphics[height=6cm, width=\textwidth, trim=0 2.4cm 0 0.6cm, clip=false]{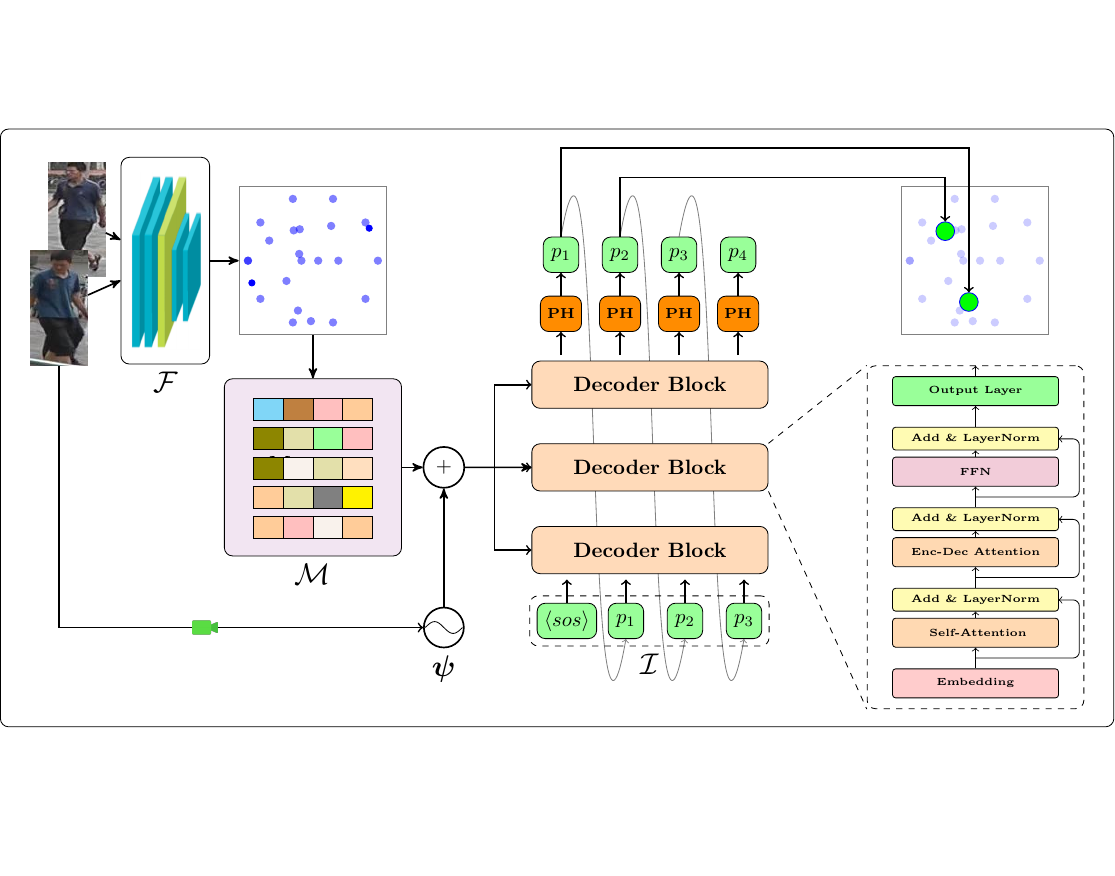}
\caption{The proposed attention-based GCP model for person Re-ID. Small blue
dots represent the extracted features $\mathcal{F}(x)$ from the backbone, while
large green dots indicate the output prototypes from the model. PH denotes the
prototype heads. For clarity, the figure displays only two of the generated
prototypes.}
\label{fig:model_1}
\end{figure}

To acquire a comprehensive representation of an individual, it is essential to
integrate subtle appearance cues observed across different images. However, not
all features are equally important. The feature integration process should
emphasize distinguishable human features while minimizing the influence of
insignificant details. To achieve this objective, we introduce a GCP model that
leverages the attention mechanism of a transformer decoder
\cite{vaswani2017attention}. Our decoder-only GCP architecture is illustrated
in Fig.~\ref{fig:model_1} and described as follows.

Given an image $x\in \mathbb{R}^{H\times W\times C}$ where $H$, $W$, and $C$
represent the height, width, and the number of channels, respectively, we first
extract the feature vector
$\mathcal{F}(x)\colon\mathbb{R}^{H\times W \times C}\to \mathbb{R}^D$
using a pretrained backbone model $\mathcal{F}$. A set of feature vectors
$\{\mathcal{F}(x_1^c),
\mathcal{F}(x_2^c), \ldots, \mathcal{F}(x_s^c)\}$ of class $c$ are combined to
encode distinct aspects of varying appearance. Camera IDs (e.g., 1, 2, 3, etc.)
associated with each image are used to generate positional embeddings
$\mathcal{\psi} \in \mathbb{R}^{S\times D}$, introducing camera and pose
information into the model. The feature vectors corresponding to different
images are treated as tokens. A variable number ($s \le |G_c|$) of tokens
ordered arbitrarily, representing the same class $c$ (i.e., individual), are
used as the memory of the decoder. Concretely, the memory fed into the GCP for
class $c$ is defined as
\begin{equation}
  \mathcal{M}^c = [\mathcal{F}(x_1^c),\mathcal{F}(x_2^c),\ldots,\mathcal{F}(x_s^c)] + \psi.
  \label{eqn:raw_input_sequence}
\end{equation}

With $\mathcal{M}^c$, the GCP model generates $N$ prototypes for class $c$ in
an autoregressive manner. We begin with an initial input sequence
$\mathcal{I}_1=[\langle sos \rangle]$ of length one, containing a learned
start-of-sequence token $\langle sos \rangle$. At each subsequent iteration
$t$, we input the sequence $\mathcal{I}_t=[\langle sos \rangle; p_1; p_2;
\ldots; p_{t-1}]$, which includes the $\langle sos \rangle$ token and all
previously generated tokens $[p_1; p_2; \ldots; p_{t-1}]$, into the decoder. We
retain only the latest generated token $p_t \in \mathbb{R}^{D}$ as the
prototype for iteration $t$. Our auto-regressive approach is essential for
creating prototypes step-by-step. Each generated prototype influences
subsequent ones, which enables the model to produce prototypes that cover
different regions within the embedding space.

Using the memory $\mathcal{M}$ and input sequence $\mathcal{I}$, the GCP model
is trained in mini-batches to minimize the distance between the prototypes
$p^c$ and feature vectors $x^c$ of the same class $c$, while simultaneously
maximizing the distance between the prototypes and feature vectors of different
classes. We realize this objective by using following loss function:
\begin{align}
  \mathcal{L}           &= \mathcal{L}_{triplet} + \lambda \mathcal{L}_{reg},\label{eqn:loss}\\
  \mathcal{L}_{triplet} &= \max(0, m + \left\Vert a - p \right\Vert_2 - \left\Vert a-n \right\Vert_2),
 \end{align}
where $m$ is the margin, and $a$, $p$, and $n$ are the anchor, positive, and
negative feature vectors, respectively. $\lambda$ serves as the constant
regularization factor. When only the triplet loss $\mathcal{L}_{triplet}$ is
used as the loss function, prototypes of the same class (i.e., $p_1^c, p_2^c,
\ldots, p_N^c$) collapse to a single point in the feature space. To prevent
this and encourage diversity, we introduce
\begin{multline}
  \mathcal{L}_{reg} =\mathbbm{1}\{k = k^\prime\} \left\Vert p_k^c -p_{k^{\prime}}^c \right\Vert_2 + \mathbbm{1}\{k \ne k^\prime\} \max(0, m -  \left\Vert p_k^c -p_{k^{\prime}}^c \right\Vert_2).
\end{multline}

%%%%%%%%%%%%%%%%%%%%%%%%%%%%%%%%%%%%%%%%%%%%%%%%%%%%%%%%%%%%%%%%%%%%%%%%%%%%%%%%
\vspace{-1mm}
\section{Evaluation}
\label{sec:evaluation}
%-------------------------------------------------------------------------------
\subsection{Datasets}
A comprehensive evaluation was performed on three commonly used person Re-ID
benchmarks: CUHK03-NP (labeled) \cite{li2014deepreid}, Market-1501
\cite{zheng2015scalable}, and MSMT17 \cite{wei2018person}.
Table~\ref{tab:dataset_statistics} summarizes the statistics of each dataset.
The CUHK03-NP dataset presents a challenging Re-ID environment due to varying
camera settings, which result in photometric transformations. Collected in
front of a supermarket, the Market-1501 dataset uses six cameras (five
high-resolution and one low-resolution). MSMT17, the largest dataset in the
evaluation, was derived from 180 hours of video footage.

\begin{table}
\centering
\begin{tabular}{|c| c| c| c| c| }
  \toprule
  Dataset     & Images  & Cameras & Train ID & Test ID \\
  \midrule
  CUHK03-NP   & 13,164  & 2       & 1,160    & 100 \\
  \midrule
  Market-1501 & 32,668  & 6       & 751      & 750 \\
  \midrule
  MSMT17      & 126,441 & 15      & 1,041    & 3,060 \\
  \bottomrule
\end{tabular}
\caption{A summary of the dataset statistics.}
\label{tab:dataset_statistics}
\vspace{-8mm}
\end{table}

%-------------------------------------------------------------------------------
\subsection{Implementation Details}
To demonstrate the robustness of our GCP model, we employed multiple backbones
to extract raw feature vectors. Akin to Luo et al. \cite{luo2019strong}, we
used a ResNet50 backbone with several optimizations to obtain global features.
Additionally, we conducted experiments using feature vectors extracted from
TransReID \cite{he2021transreid} and PHA \cite{zhang2023pha}. We extracted
features from the model pretrained on the training split of the dataset. These
extracted features serve as memory $\mathcal{M}$ for the GCP model from which
the final class prototype vectors $p_i$ are derived (Fig.~\ref{fig:model_1}).

GCP comprises six individual decoder blocks, each with multi-headed
self-attention to process inputs and multi-headed cross-attention to attend to
memory based on the input. Each layer includes four attention heads, with a
feed-forward network dimension of 512 and a dropout rate set to 0.2.  Prototype
heads PH, positioned above the decoder, share the same parameters. A PH
consists of a single perceptron layer ($\mathbb{R}^n \to \mathbb{R}^n$), where
$n$ represents the feature dimension.  To identify the optimal number of
prototypes for the GCP model, we empirically evaluated different numbers ($N$)
of prototypes. Although we generate $N$ prototypes (tokens) at the output
layer, our model is adaptable to any number of prototypes due to its
autoregressive capability.

The GCP model was trained end-to-end using the loss function $\mathcal{L}$
defined by \eqref{eqn:loss}. In the experiments, $m$ was set to 1.2 and the
network was trained using stochastic gradient descent with a learning rate of
0.01, momentum equal to 0.9, and a weight decay of $5\times10^{-4}$. We used a
feature dimension size of 2048 for the ResNet50 backbone and 3840 for TransReID
and PHA backbones with a batch size of 128 (16 classes with 8 instances per
class). In addition, the training images were augmented with horizontal flip
and normalization.

\begin{figure}
\centering
 \begin{tikzpicture}[scale=0.9,
    vec/.style={ draw, rounded corners, rectangle, minimum height={21pt}, minimum width={7}, font=\small},
    arr/.style = {draw, thick, -latex',shorten >=2pt}]
        %inital feature set
  \definecolor{ashgrey}{rgb}{0.7, 0.75, 0.71}
  %\node[rectangle, rounded corners, draw = gray, text = olive, fill =
   %ashgrey!60, minimum width = 5cm, minimum height = 4.6cm] (r) at (2.3,.9) {};

    \node[vec, fill=red!30]at (0, 1.7) (a1) {\textcolor{white}{}};
    \node[vec, right = 0cm of a1.east,fill=red!30] (a2)  {\textcolor{white}{}};
    \node[vec, right = 0cm of a2.east,fill=black!30!green!30, inner sep=0] (b1)  {\textcolor{black}{$\kappa$}};
    \node[vec, right = 0cm of b1.east,fill=black!30!green!30, inner sep=0] (b2)  {\textcolor{black}{$\kappa$}};
    \node[vec, right = 0cm of b2.east,fill=black!30!blue!30] (c)  {\textcolor{white}{}};
    \draw [decorate,decoration={brace,amplitude=5pt,raise=4ex}](-.2,1.7) -- (1.3,1.7) node[midway,yshift=3em](x){\small Gallery};

    %filter
    \node[ right = .5cm of c,fill=purple!30!blue!50,draw,rounded corners] (filter)  {\textcolor{white}{\small Filter}};

    %final feature se
    \node[vec, right = .8cm of filter.east, fill=red!30] (fa1) {\textcolor{white}{}};
    \node[vec, right = 0cm of fa1.east, fill=red!30] (fa2)  {\textcolor{white}{}};
    \node[vec, right = 0cm of fa2.east, fill=black!30!blue!30] (fc)  {\textcolor{white}{}};

    %gcp
    \node[ right = .8cm of fc.east,fill=purple!30!blue!50,draw,rounded corners] (gcp)  {\textcolor{white}{\small GCP}};

    %final prototype
    \node[vec, anchor=east, fill=yellow!40,draw,rounded corners, ] at (
    $(gcp.east)+(1,0)$) (proto) {\textcolor{white}{}};
    \node[vec,  anchor=west, fill=yellow!40,draw,rounded corners, ] at (
    $(proto.east)+(0,0)$)(proto1) {\textcolor{white}{}};
    \draw [decorate,decoration={brace,amplitude=5pt,raise=4ex}](7.2,1.6)
    -- +(.6,0) node[midway,yshift=3em](x){\small Prototype};

    %query
    \node[inner sep=0,  above= .3cm of filter.north] (qc) {\includegraphics[width=12pt, ]{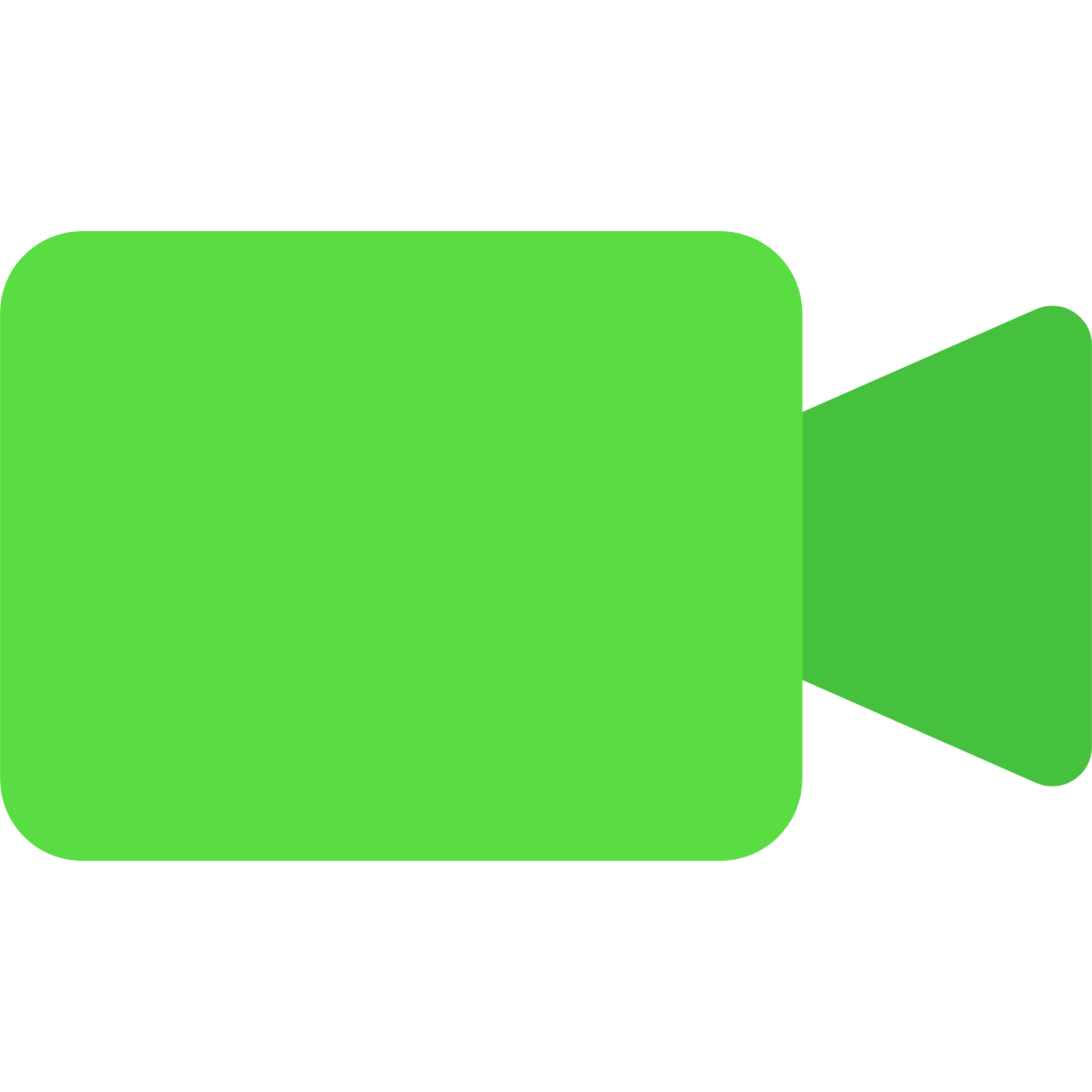}};
    \node[above = .2cm of qc, anchor=north,minimum height = 8pt, minimum width = 25pt, fill=black!30!green!30, draw,rounded corners, label={[name=qal]right:\small Query}, inner sep=0] (qa)  {\textcolor{black}{$\kappa$}};

    % arrows
    \draw[arr] (c) --  (filter);
    \draw[arr] (filter) --  (fa1);
    \draw[arr] (fc) --  (gcp);
    \draw[arr] (gcp) --  (proto);
    \draw[arr] (qc) --  (filter);

%    \node[ rectangle, rounded corners, draw = gray, text = olive, fill = ashgrey!60 ,fill opacity=0.4,fit=(a1)(proto)(qc)(x)(qal)(protol) ] {};
\end{tikzpicture}
\caption{Prototype generation during inference via the proposed GCP model. Each
ellipse signifies a feature vector, with colors (e.g., red, green, blue)
indicating the associated camera ID. When dealing with a particular query
captured by a green camera, feature vectors marked in green are excluded during
prototype generation.}
\label{fig:inference_camera_filter}
\vspace{-4mm}
\end{figure}
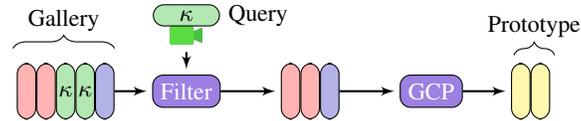

During the inference stage, given a query $q$ with class $c$ and camera
$\kappa$, we excluded all gallery instances from the same $c$ that were captured
using the same $\kappa$ while generating gallery prototypes for $c$. This
process is illustrated in Fig.~\ref{fig:inference_camera_filter}. For prototypes
of other classes, we used all available gallery features regardless of their
cameras. We generated $N$ prototypes sequentially in $N$ iterations. The model
was trained for 120 epochs on an Ubuntu 18.04 LTS machine with an Intel i7-8700
CPU, 64 GB of RAM, and an NVIDIA A100 GPU.

%-------------------------------------------------------------------------------
\subsection{Experiments}
The performance of our approach was measured using two metrics: cumulative
matching characteristic at rank-1 (R-1) and mean average precision (mAP).
Evaluations were conducted in a single-query setting without re-ranking. The
GCP model demonstrated significant improvement in R-1/mAP across all base
feature extractor models. Notably, with features extracted from the PHA model,
the GCP model achieved  93.1\%/92.2\%, 97.3\%/97.1\%, and 89.7\%/83.4\% in
R-1/mAP on the CUHK03-NP, Market-1501, and MSMT17 datasets, respectively.

We also compared against two baseline methods capable of generating $N$
prototypes per class from the gallery set. The first and simpler approach is a
clustering-based prototype selection algorithm, referred to as
\textit{k-centroid}. In this method, we applied k-means clustering to identify
$N$ clusters within each class and selected the cluster centroids as
representative prototypes. The second method is based on farthest-point
sampling (FPS) \cite{eldar1997farthest,kamousi2016analysis}. FPS involves
iteratively selecting points that are maximally distant from previously
selected ones, thereby producing a subset that effectively approximates the
diversity of the original distribution with fewer points.

In FPS, each feature vector $\mathcal{F}(x)\in R^n$ is treated as a point.
However, the standard FPS technique may select prototypes located at the class
boundaries resulting in a poor mAP. To address this issue, we created a
modified version called $\alpha$-farthest point sampling ($\alpha$-FPS), which
is outlined in Algorithm~\ref{alg:afps_algorithm}. Additional details on the
$\alpha$-FPS algorithm are provided in the appendix.

\begin{algorithm}
\caption{$\alpha$-Farthest Point Sampling}
\label{alg:afps_algorithm}
\hspace*{\algorithmicindent} \textbf{Input: $X, N, \alpha$}\\
\hspace*{\algorithmicindent} \textbf{Output: $P$}
\begin{algorithmic}[1]
%\Require $n \geq 1$
%\Ensure $y = x^n$
\State $P \gets \{centroid(X)\}$ \label{lst:line:init}
%\State $X \gets x$
\While{$N \ge 1$}
  \State $x,p \gets farthest(X,P)$ \label{lst:line:fps}
  \State $X \gets X \setminus x$ \label{lst:line:remove}
  \State $x \gets x+\alpha(p-x)$ \label{lst:line:modify}
  \State $P \gets P \cup \{x\}$ \label{lst:line:add}
  \State $N \gets N - 1$
\EndWhile
\end{algorithmic}
\end{algorithm}

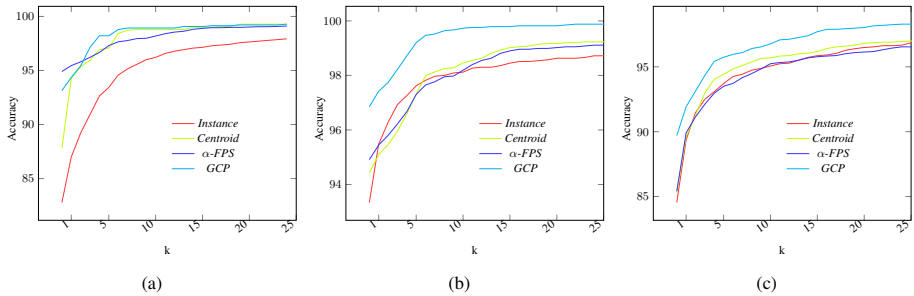
\begin{figure}
%\centering
%\hspace{-7pt}
\begin{subfigure}{0.33\textwidth}
\centering
\pgfplotstableread[col sep=comma,]{images/plots/cuhk03_acc.csv}\datatable
\begin{tikzpicture}[scale=0.5]
  \begin{axis}[
    legend style={draw=none, font=\small, at={(0.5,.5)}, anchor=north west,legend columns=1},
    xtick=data,
    %xticklabels from table={\datatable}{x-axis},
    y label style={at={(axis description cs:-0.1,0.35)},anchor=west},
    x label style={at={(axis description cs:0.5,-0.1)},anchor=north},
    xtick={1,5,10,15,20,25,30,40,50},
    xticklabels={1,5,10,15,20,25,30,40,50},
    xmax=25,
    label style={font=\footnotesize},
    x tick label style={{font=\footnotesize}, rotate=35, anchor=east},
    y tick label style={{font=\footnotesize}},
    legend style={{font=\footnotesize}},
    ylabel={Accuracy},
    xlabel={k}]

    \addplot [mark=., red!80 ] table [x expr=\coordindex, y={base}]{\datatable};
    \addlegendentry{\emph{Instance}}

    \addplot [mark=., lime ] table [x expr=\coordindex, y={centroid}]{\datatable};
    \addlegendentry{\emph{Centroid}}

    \addplot [mark=., blue!80] table [x expr=\coordindex, y={alpha}]{\datatable};
    \addlegendentry{$\alpha$-\emph{FPS}}

    \addplot [mark=., cyan ] table [x expr=\coordindex, y={trans}]{\datatable};
    \addlegendentry{\emph{GCP}}
  \end{axis}
\end{tikzpicture}
\caption{}
\label{fig:topk_cuhk03}
\end{subfigure}
\begin{subfigure}{0.33\textwidth}
\centering
\pgfplotstableread[col sep=comma,]{images/plots/market_acc.csv}\datatable
\begin{tikzpicture}[scale=0.5]
  \begin{axis}[
    legend style={draw=none, font=\small, at={(0.5,.5)}, anchor=north west,legend columns=1},
    xtick=data,
    %xticklabels from table={\datatable}{x-axis},
    y label style={at={(axis description cs:-0.1,0.35)},anchor=west},
    x label style={at={(axis description cs:0.5,-0.1)},anchor=north},
    xtick={1,5,10,15,20,25,30,40,50},
    xticklabels={1,5,10,15,20,25,30,40,50},
    xmax=25,
    label style={font=\footnotesize},
    x tick label style={{font=\footnotesize}, rotate=35, anchor=east},
    y tick label style={{font=\footnotesize}},
    legend style={{font=\footnotesize}},
    ylabel={Accuracy},
    xlabel={k}]

    \addplot [mark=., red!80 ] table [x expr=\coordindex, y={base}]{\datatable};
    \addlegendentry{\emph{Instance}}

    \addplot [mark=., lime ] table [x expr=\coordindex, y={centroid}]{\datatable};
    \addlegendentry{\emph{Centroid}}

    \addplot [mark=., blue!80] table [x expr=\coordindex, y={alpha}]{\datatable};
    \addlegendentry{$\alpha$-\emph{FPS}}

    \addplot [mark=., cyan ] table [x expr=\coordindex, y={trans}]{\datatable};
    \addlegendentry{\emph{GCP}}
  \end{axis}
\end{tikzpicture}
\caption{}
\label{fig:topk_market}
\end{subfigure}
\begin{subfigure}{0.33\textwidth}
\centering
\pgfplotstableread[col sep=comma,]{images/plots/msmt17_acc.csv}\datatable
\begin{tikzpicture}[scale=0.5]
  \begin{axis}[
    legend style={draw=none, font=\small, at={(0.5,.5)}, anchor=north west,legend columns=1},
    xtick=data,
    %xticklabels from table={\datatable}{x-axis},
    x label style={at={(axis description cs:.5,-0.1)},anchor=north},
    y label style={at={(axis description cs:-0.1,0.35)},anchor=west},
    xtick={1,5,10,15,20,25,30,40,50},
    xticklabels={1,5,10,15,20,25,30,40,50},
    xmax=25,
    label style={font=\footnotesize},
    x tick label style={{font=\footnotesize}, rotate=35, anchor=east},
    y tick label style={{font=\footnotesize}},
    legend style={{font=\footnotesize}},
    ylabel={Accuracy},
    xlabel={k}]

    \addplot [mark=., red!80 ] table [x expr=\coordindex, y={base}]{\datatable};
    \addlegendentry{\emph{Instance}}

    \addplot [mark=., lime] table [x expr=\coordindex, y={centroid}]{\datatable};
    \addlegendentry{\emph{Centroid}}

    \addplot [mark=., blue!80] table [x expr=\coordindex, y={alpha}]{\datatable};
    \addlegendentry{$\alpha$-\emph{FPS}}

    \addplot [mark=., cyan!90 ] table [x expr=\coordindex, y={trans}]{\datatable};
    \addlegendentry{\emph{GCP}}
  \end{axis}
\end{tikzpicture}
\caption{}
\end{subfigure}
\caption{Top-$k$ accuracy on the (a) CUHK03-NP, (b) Market-1501, and (c) MSMT17
datasets.}
\label{fig:top-k_accuracy}
\end{figure}

%------------------------------------------------------------------------------%
\noindent\textbf{Quantitative evaluation.} Table~\ref{tab:sota} provides a
comparison of our results with both state-of-the-art Re-ID methods and baseline
methods. We experimented with different values of $N$ for the baseline methods
and reported the best-performing results. On mAP, the GCP model surpasses most
existing person Re-ID algorithms across all datasets. Notably, the
3840-dimensional feature vector extracted from the TransReID and PHA backbones
exhibited superior performance compared to the 2048-dimensional feature vector
generated by the ResNet50 backbone.

\begin{table}
\setlength\tabcolsep{10pt}
\centering
\begin{tabular}{l| c c c c c c  }
\toprule
\multirow{2}*{Method} & \multicolumn{2}{c}{CUHK03-NP} & \multicolumn{2}{c}{Market-1501}& \multicolumn{2}{c}{MSMT17} \\
\cmidrule(lr){2-3}
\cmidrule(lr){4-5}
\cmidrule(lr){6-7}
 & R-1 & mAP & R-1 & mAP & R-1 & mAP\\ \midrule
%PCB+RPP (ECCV'18)\cite{sun2018beyond} & 93.8 & 81.6 & 83.3 & 69.2&68.2&40.4&-&-&63.\\
%GCS (CVPR'18)\cite{chen2018group} &93.5&81.6&84.9 & 69.5&-&-&-&-&-\\
  OSNet (ICCV'19)\cite{zhou2019omni}                        & -             & -               & 94.8          & 84.9                 & 78.7         & 52.9\\
  Pyramid (CVPR'19)\cite{zheng2019pyramidal}                & 78.9          & 76.9            & 95.7          & 88.2                 & -            & -\\
  ABDNet (CVPR'19) \cite{chen2019abd}                       & -             & -               & 88.3          & 95.6                 & 60.8         & 82.3\\
  CBDB-Net (TCSVT'21)\cite{tan2021incomplete}               & 77.8          & 76.6            & 94.4          & 85.0                 & -            & -\\
  Auto-ReID (CVPR'19) \cite{quan2019auto}                   & 77.9          & 73.0            & 95.7          & 88.2                 & -            & -\\
  st-ReID (AAAI'19)\cite{wang2019spatial}                   & -             & -               & 94.5          & 85.1                 & -            & -\\
  C2F (CVPR'21)\cite{zhang2021coarse}                       & 80.6          & 79.3            & 94.8          & 87.7                 & -            & -\\
  CDNet (CVPR'21) \cite{li2021combined}                     & -             & -               & 94.8          & 87.7                 & 78.9         & 54.7\\
  PAT (CVPR'21)\cite{li2021diverse}                         & -             & -               & 95.4          & 88.0                 & -            & -\\
  NFormer (CVPR'22)\cite{wang2022nformer}                   & 78.0          & 77.2            & 94.7          & 91.1                 & 77.3         & 59.8\\
  BPBreID\textsubscript{HR} (WACV'23)\cite{somers2023body}  & -             & -               & 95.7          & 93.0                 & -            & -\\
  SOLIDER (CVPR'23)\cite{chen2023beyond}                    & -             & -               & 95.7          & 89.4                 &\textbf{90.7} & 77.1\\
  TransReID (CVPR'21)\cite{he2021transreid}                 & 81.7          & 79.6            & 88.9          & 95.2                 & 85.3         & 67.4\\
  SP loss (CVPR'23)\cite{zhou2023adaptive}                  & 82.4          & 84.6            & 89.6          & 80.5                 & 82.3         & 61.0\\
  PHA (CVPR'23)\cite{zhang2023pha}                          & 84.5          & 83.0            & 96.1          & 90.2                 & 86.1         & 68.9\\
  IRM (CVPR'24)\cite{he2024instruct}                        & 86.5          & 85.4            & 96.5          & 93.5                 & 86.9         & 72.4\\

\midrule
\midrule
  k-centroid + (PHA)                                        & 85.9          & 84.2            & 94.4          & 94.8                 & 85.1         & 81.1\\
  $\alpha$-FPS + (PHA)                                      & 86.8          & 84.8            & 95.9          & 95.8                 & 85.4         & 81.5\\
  GCP (ours) + (ResNet50)                                   & 88.6          & 88.1            & 89.5          & 95.4                 & -            & -\\
  GCP (ours) + (TransReID)                                  & -             & -               & 95.3          & 95.9                 & 85.4         & 82.1\\
  GCP (ours) + (PHA)                                        & \textbf{93.1} & \textbf{92.2}   & \textbf{97.3} & \textbf{97.1}        & 89.7         & \textbf{83.4}\\
\bottomrule
\end{tabular}
\caption{Quantitative results on the CUHK03-NP, Market-1501, and MSMT17
datasets. R-1 is top-1 accuracy and mAP is mean average precision. The best
performance value for each column is marked in bold.}
\label{tab:sota}
\end{table}

To ensure a fair comparison, we evaluated the GCP model alongside other
prototype-based methods, including our baseline methods. We calculated top-$k$
accuracy for $k\in\{1, 2, \ldots, 25\}$, with the results displayed in
Fig.~\ref{fig:top-k_accuracy}. These results show that the GCP approach
consistently outperforms other methods in top-$k$ accuracy across all $k$
values.

\begin{figure}
\centering
\resizebox{\linewidth}{!}{%\usepackage{xcolor}
%\usepackage{pgfplots}
%\usepackage{tikz}

% Define bar chart colors
%
\definecolor{bblue}{HTML}{4F81BD}
\definecolor{rred}{HTML}{85c1e9}
\definecolor{ggreen}{HTML}{27ae60}
\definecolor{ppurple}{HTML}{884ea0}

\begin{tikzpicture}
    \begin{axis}[
        title = CUHK03-NP,
        width  = 0.35*\textwidth,
        height = 6cm,
        %major x tick style = transparent,
        ticklabel style = {font=\large},
        ybar=2*\pgflinewidth,
        bar width = 5pt,
        ymajorgrids = true,
        ylabel = {Frequency},
        xlabel = {Images per person},
        xtick = {2,4,6,8,10},
        scaled y ticks = false,
        ymin=0,
        legend cell align=left,
        legend style={
                at={(.8,.75)},
                anchor=south east,
                column sep=1ex
        },
       %x tick label style={rotate=45},
       %y label style={font=\Large},
       extra x tick style={grid=major,},
    ]
       \addplot[ybar,fill=cyan] table [x, y, col sep=comma] {images/plots/cuhk03_hist.csv};
    \end{axis}
\end{tikzpicture}
\begin{tikzpicture}
    \begin{axis}[
        title = Market-1501,
        width  = 0.5*\textwidth,
        height = 6cm,
        %major x tick style = transparent,
        ticklabel style = {font=\large},
        ybar=2*\pgflinewidth,
        bar width = 5pt,
        ymajorgrids = true,
        ylabel = {Frequency},
        xlabel = {Images per person},
        xtick={10,20,30,40,50,60},
        scaled y ticks = false,
        ymin=0,
        legend cell align=left,
        legend style={
                at={(.98,.75)},
                anchor=south east,
                column sep=1ex
        },
        %x tick label style={rotate=45},
        %y label style={font=\Large},
        extra x tick style={grid=major,},
    ]
       \addplot[ybar,fill=cyan] table [x, y, col sep=comma] {images/plots/market_hist.csv};

%        \addplot[style={ppurple,fill=ppurple,mark=none}]
%             coordinates {(Tree01,0.74) (Tree 02,1.07) (Tree03,1.23)};
    \end{axis}
\end{tikzpicture}
\begin{tikzpicture}
    \begin{axis}[
        title = MSMT17,
        width  = 0.5*\textwidth,
        height = 6cm,
        %major x tick style = transparent,
        ticklabel style = {font=\large},
        ybar=2*\pgflinewidth,
        bar width=5pt,
        ymajorgrids = true,
        ylabel = {Frequency},
        xlabel = {Images per person},
        xtick = {25,50,100,150,200},
        scaled y ticks = false,
        ymin=0,
        legend cell align=left,
        legend style={
                at={(.8,.75)},
                anchor=south east,
                column sep=1ex
        },
        %x tick label style={rotate=45},
        %y label style={font=\Large},
extra x tick style={grid=major,},
    ]
       \addplot[ybar,fill=cyan] table [x, y, col sep=comma] {images/plots/msmt_hist.csv};
    \end{axis}
\end{tikzpicture}}
\caption{A histogram of the number of images per person in the gallery set.}
\label{fig:group_evaluation_hist}
\end{figure}
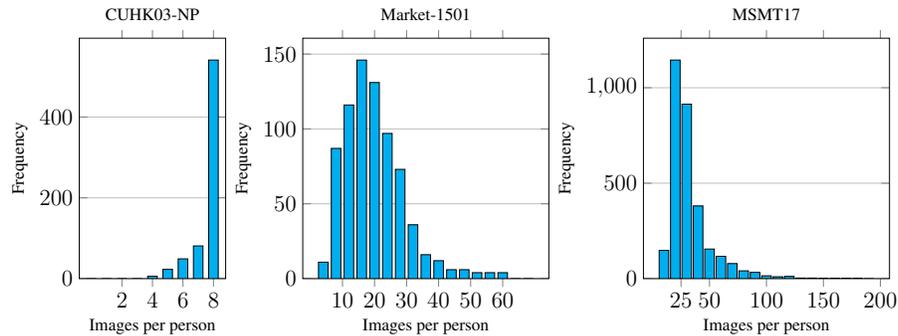

\setlength{\tabcolsep}{4pt}
\definecolor{apple green}{rgb}{0.55, 0.71, 0.0} %apple green
\begin{table}
\centering
\begin{adjustbox}{max width=\linewidth}
\begin{tabular}{l | c c c c|  c c c c| c c c c}
\toprule
Dataset& \multicolumn{4}{c}{CUHK03-NP} & \multicolumn{4}{c}{Market-1501} & \multicolumn{4}{c}{MSMST17} \\
\cmidrule(lr){1-1}
\cmidrule(lr){2-5}
\cmidrule(lr){6-9}
\cmidrule(lr){10-13}
Subset & 4-5 & 6 & 7 & 8 & 1-15 & 16-30 & 31-50 & 50+ & 1-10 & 11-20 & 21-30 & 30+ \\ \midrule
mAP    & 91.9 & 92.1 & 92.2 & 92.3 & 80.2 & 84.4 & 83.3 & 80.5 & 96.1 & 97.5 & 97.2 & 97.1 \\
\bottomrule
\end{tabular}
\end{adjustbox}
\caption{The mAP for different subsets of the CUHK03-NP, Market-1501, and
MSMT17 datasets.}
\label{tab:group_evaluation}
\vspace{-2mm}
\end{table}

The number of images per person available in the gallery set varies as depicted
in Fig.~\ref{fig:group_evaluation_hist}.  Since GCP derives prototypes by
aggregating features from multiple images of the same class, the number of
images used in the prototype formation is expected to impact performance
substantially. To assess this effect, we segmented the gallery set into
distinct groups based on the number of images per class and evaluated model
performance within each group.  The results presented in
Table~\ref{tab:group_evaluation} demonstrate that optimal performance is
attained with 8 gallery images for the CUHK03-NP dataset, 10–20 gallery images
for the Market-1501 dataset, and 16–30 gallery images for the MSMT17 dataset.
Given the greater diversity and scale of the MSMT17 dataset, the model requires
a larger number of images to construct a comprehensive representation of an
individual compared to the Market-1501 dataset.

\begin{figure}
\centering
\begin{subfigure}{0.35\linewidth}
  \centering
  \includegraphics[width=\linewidth]{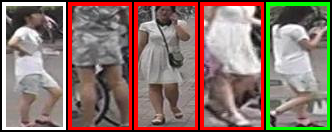}
\end{subfigure}\quad
\begin{subfigure}{0.35\linewidth}
  \centering
  \includegraphics[width=\linewidth]{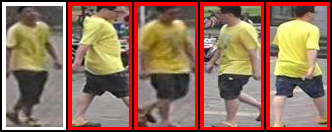}
\end{subfigure}%

\begin{subfigure}{0.35\linewidth}
  \centering
  \includegraphics[width=\linewidth]{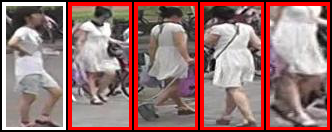}
\end{subfigure}\quad
\begin{subfigure}{0.35\linewidth}
  \centering
  \includegraphics[width=\linewidth]{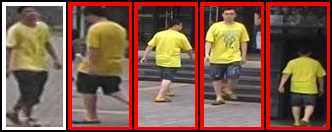}
\end{subfigure}

\begin{subfigure}{0.35\linewidth}
  \centering
  \includegraphics[width=\linewidth]{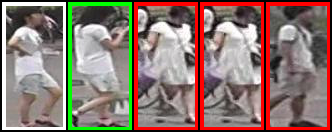}
\end{subfigure}\quad
\begin{subfigure}{0.35\linewidth}
  \centering
  \includegraphics[width=\linewidth]{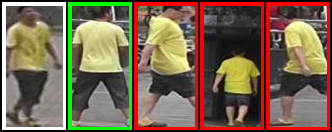}
\end{subfigure}\quad

\begin{subfigure}{0.35\linewidth}
  \centering
  \includegraphics[width=\linewidth]{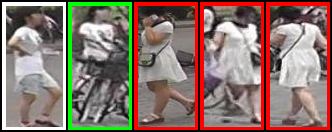}
\end{subfigure}\quad
\begin{subfigure}{0.35\linewidth}
\centering
  \includegraphics[width=\linewidth]{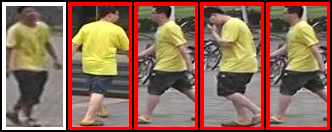}
\end{subfigure}
\caption{Examples of retrieved person Re-ID images. The first two rows show the
results of instance-based and centroid-based methods, respectively. The images
in the third and fourth rows are the results of our attention-based GCP and
$\alpha$-FPS methods, respectively. In each row, the query images are enclosed
by white borders. Images to the right of the query display the top four
retrieved images. The green/red borders indicate whether the images share the
same/different identities as the query.}
\label{fig:qualitative_results}
\vspace{-6mm}
\end{figure}

%------------------------------------------------------------------------------%
\noindent\textbf{Qualitative evaluation.} In
Fig.~\ref{fig:qualitative_results}, we display the top four retrieved images
for queries using different methods. For GCP, the retrieved images correspond
to those closest to the prototypes in the feature space. Both the people in
white and yellow shirts are successfully identified at R-1 by the GCP model
(third row). The fourth row illustrates the limitations of the baseline
$\alpha$-FPS algorithm, e.g., where it fails to retrieve the person with the
correct identity. Nevertheless, $\alpha$-FPS performs better than the
instance-based and centroid-based methods in the case of identifying the person
in the white shirt.

\begin{figure}
\centering
\input{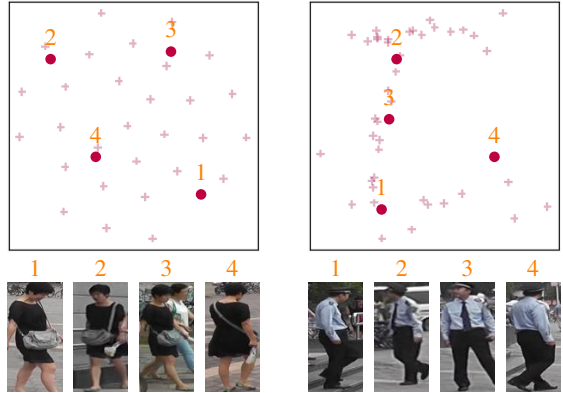}
\caption{Prototypes generated using GCPs with their nearest gallery images in
the feature space. The number above each prototype indicates the $i^{th}$
iteration in which it was generated.}
\label{fig:embedding_qualitative}
\vspace{-2mm}
\end{figure}

To qualitatively examine the use of GCPs, we selected all gallery images of an
arbitrary class to generate $N$ prototypes in sequence. Since each prototype
represents only a point in feature space, we tag each with its nearest gallery
image in the feature space to provide a visual reference. The generated
prototypes and their corresponding tagged images are shown in
Fig.~\ref{fig:embedding_qualitative}. The GCP approach is able to effectively
identify key images that cover the feature space.

%-------------------------------------------------------------------------------
\subsection{Discussion}
The number of prototypes is a critical parameter in the GCP model. We conducted
several experiments that vary the number of prototypes per class. As shown in
Table~\ref{tab:results_varying_N}, increasing the number of prototypes per
class enhances accuracy, but decreases mAP. This is because a higher number of
prototypes shifts the approach towards instance-based Re-ID. Unless stated
otherwise, all results reported in this paper use $N=3$.

\begin{table}
\centering
\setlength\extrarowheight{-2pt}
\begin{tabularx}{.85\columnwidth}{Y|YY YY}
\hline
\multirow{2}*{$N$} & \multicolumn{2}{c}{CUHK03-NP} & \multicolumn{2}{c}{Market-1501}\\
\cmidrule(lr){2-3}
\cmidrule(lr){4-5}
&R-1 & mAP &R-1 & mAP\\
\hline
2       & 92.2          & 92.2          & 97.1            & \textbf{97.2}\\
3       & 93.1          & \textbf{92.2}       & 97.3            & 97.1         \\
4       & 93.5          & 91.9          & 97.3            & 97.0         \\
5       & 94.0          & 91.7         & \textbf{97.5}   & 97.0         \\
6       & \textbf{94.3} & 91.4          & 97.5            & 96.8         \\ \hline
\end{tabularx}
\caption{The effect of the number of prototypes ($N$) per class on the
performance using the CUHK03-NP and Market-1501 datasets.}
\label{tab:results_varying_N}
\vspace{-2mm}
\end{table}

It is worth noting that during the inference stage, the class information of
the gallery images is utilized to group them and generate prototypes. Exposure
to such information is common in practical applications. For example, in
surveillance scenarios, multiple images of a target individual are often known
in advance and readily available for comparison against the query image.

%%%%%%%%%%%%%%%%%%%%%%%%%%%%%%%%%%%%%%%%%%%%%%%%%%%%%%%%%%%%%%%%%%%%%%%%%%%%%%%%
\vspace{-1mm}
\section{Conclusion}
\label{sec:conclusion}
This paper introduced the concept of a GCP, which encapsulates both
instance-based and centroid-based image retrieval. To do this, we first
analyzed the intricacies of various types of class representatives. Then, we
introduced a learning-based architecture to generate robust class prototypes.
Additionally, we developed a straightforward yet effective algorithm,
$\alpha$-FPS, as a baseline method for selecting prototypes without requiring
model training. Experimental results show that our GCP approach surpasses
modern techniques on person Re-ID benchmark datasets. In future work, we aim to
further refine this methodology to identify superior class prototypes within
the generalized framework.

%%%%%%%%%%%%%%%%%%%%%%%%%%%%%%%%%%%%%%%%%%%%%%%%%%%%%%%%%%%%%%%%%%%%%%%%%%%%%%%%
\vspace{-1mm}
\section*{Acknowledgments}
This material is based upon work supported by the Air Force Research Laboratory
under award number FA8571-23-C-0041. 

\bibliographystyle{splncs04}
\bibliography{references}

\section*{Appendix}
\appendix
In this appendix we provide additional details on the GCP methodology. 

\vspace{-1mm}
\section{Rationale Behind the Effectiveness of GCP}
\begin{figure}
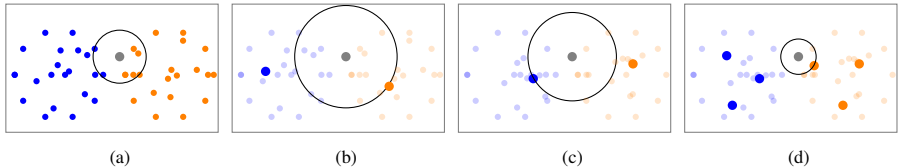

\centering
\begin{subfigure}{0.25\textwidth}
  %\centering
  \pgfmathsetseed{24}% 
  \includegraphics[width=0.93\textwidth]{images/individual_proxy.tikz}
  \caption{}
  \label{fig:one_to_one_proxy}
\end{subfigure}
\hspace{-2mm}
\begin{subfigure}{0.25\textwidth}
  %\centering
  \pgfmathsetseed{24}%
  \includegraphics[width=0.93\textwidth]{images/centroid_ok.tikz}
  \caption{}
  \label{fig:centroid_proxy}
\end{subfigure}
\hspace{-2mm}
\begin{subfigure}{0.25\textwidth}
  %\centering
  \pgfmathsetseed{2}%
  \includegraphics[width=0.93\textwidth]{images/centroid_error.tikz}
  \caption{}
  \label{fig:centroid_error}
\end{subfigure}
\hspace{-2mm}
\begin{subfigure}{0.25\textwidth}
  %\centering
  \pgfmathsetseed{2}%
  \includegraphics[width=0.93\textwidth]{images/multi_proxy.tikz}
  \caption{}
  \label{fig:multi_prototype}
\end{subfigure}
\caption{A comparison between (a) instance-based, (b, c) centroid-based, and (d)
prototype-based Re-ID.}
\label{fig:reid_comparison}
\vspace{-4mm}
\end{figure}

To demonstrate how GCP enhances Re-ID performance, we present a toy example.
Consider the distribution of feature vectors for two neighboring classes in the
embedding space as depicted in Fig.~\ref{fig:reid_comparison}. In
Fig.~\ref{fig:one_to_one_proxy}, the gray dot denotes the query feature vector,
whose true class label is orange. If we compute the precision of this query
image using the five nearest neighbors, the precision will be $0.6$ (three true
positives and two false positives). Conversely, if we employ a centroid
prototype to represent the class instead of using all the individual feature
vectors, we can achieve a precision of 1.0 (Fig.~\ref{fig:centroid_proxy}).
Thus, using the centroid prototype can improve the precision for certain
distributions of gallery features.

However, centroid prototypes are not robust to arbitrary image distributions.
Consider a different class distribution as illustrated in
Fig.~\ref{fig:centroid_error}. Unlike the scenario in
Fig.~\ref{fig:centroid_proxy}, many gallery images of the blue class are closer
to the orange class. Hence, the blue class centroid is shifted towards the
orange class. For such a distribution, if we use the centroid prototype as the
class representative and calculate precision based on it, the precision will be
very low and the accuracy will be reduced as well. This example underscores the
limitation of using a centroid vector as a class representative, it is not
robust to arbitrary image distributions. Our GCP method addresses this
impediment by considering the gallery image distribution,
Fig.~\ref{fig:multi_prototype}.

\section{$\alpha$-Farthest Point Sampling}
The input to Algorithm~\ref{alg:afps_algorithm} is the set $X$ of class feature
vectors, the number $N$ of prototypes to select, and the parameter $\alpha$.
By tuning $\alpha$ we can regulate the degree of interpolation between the
selected farthest point and the prior prototype. The output of the algorithm is
a set of prototypes $P$ representing the given class. Concretely, the algorithm
starts with a single prototype, which is the centroid of the given feature
vectors (line~\ref{lst:line:init}). Next, it iteratively selects a point $x \in
X$ that is farthest from the currently selected set of prototypes $P$, followed
by a $p \in P$ that is closest to $x$ (line~\ref{lst:line:fps}). Subsequently,
$x$ is removed from the available set of features $X$
(line~\ref{lst:line:remove}). The algorithm then modifies $x$ based on the
given $\alpha$ value (line~\ref{lst:line:modify}) and adds it to the set of
prototypes $P$ (line~\ref{lst:line:add}).

%The input to Algorithm~{\color{red}{1}} is the set $X$ of class feature
%vectors, the number $N$ of prototypes to select, and the parameter $\alpha$.
%By tuning $\alpha$ we can regulate the degree of interpolation between the
%selected farthest point and the prior prototype. The output of the algorithm is
%a set of prototypes $P$ representing the given class. Concretely, the algorithm
%starts with a single prototype, which is the centroid of the given feature
%vectors (line~{\color{red}{1}}). Next, it iteratively selects a point $x \in X$
%that is farthest from the currently selected set of prototypes $P$, followed by
%a $p \in P$ that is closest to $x$ (line~{\color{red}{3}}). Subsequently, $x$
%is removed from the available set of features $X$ (line~{\color{red}{4}}).
%The algorithm then modifies $x$ based on the given $\alpha$ value
%(line~{\color{red}{5}}) and adds it to the set of prototypes $P$
%(line~{\color{red}{6}}).

\vspace{-1mm}
\section{Convergence Analysis}
\vspace{-6mm}
\begin{figure*}[ht!]
\centering
\begin{subfigure}{0.32\textwidth}
  \begin{framed}
    \includegraphics[width=.9\linewidth, height= .65\linewidth]
    {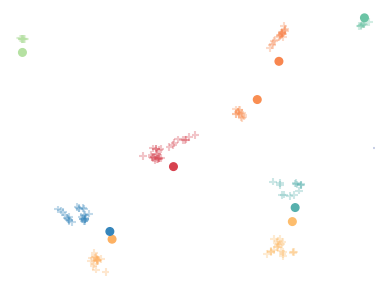}
  \end{framed}
\caption{}
\end{subfigure}
\begin{subfigure}{0.32\textwidth}
  \begin{framed}
    \includegraphics[width=.9\linewidth, height= .65\linewidth]{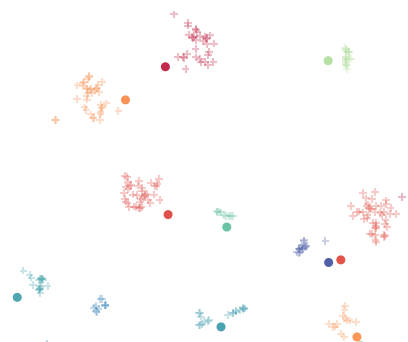}
  \end{framed}
\caption{}
\end{subfigure}
\begin{subfigure}{0.32\textwidth}
  \begin{framed}
    \includegraphics[width=.9\linewidth, height= .65\linewidth ]{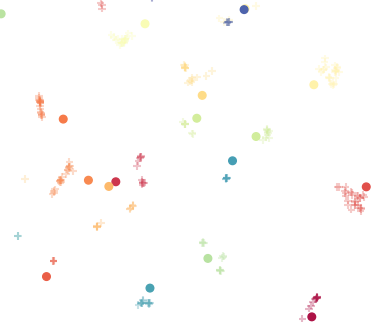}
  \end{framed}
\caption{}
\end{subfigure}
\caption{A subset of the feature vectors (+) in the embedding space for the (a)
CUHK03-NP, (b) Market-1501, and (c) MSMT17 datasets. The colored dots represent
the class prototypes selected by the GCP method. Best viewed zoomed in.}
\label{fig:gallery_vs_centroid}
\vspace{-4mm}
\end{figure*}

We sought to investigate whether GCP converges to a centroid-based approach
when $N=1$, i.e., when only a single prototype is generated per class.
Specifically, we aimed to determine whether the prototype produced by GCP
closely resembles the centroid of the class in terms of proximity. To explore
this, we conducted multiple experiments across different datasets.
Interestingly, GCP did not select the centroid as the prototype in most cases.
Instead, it generated prototypes influenced by the distribution of images in
the embedding space, which were often located far from the class centroids. To
illustrate this, we visualized the feature vectors in a 2D space using t-SNE
\cite{van2008visualizing} as shown in Fig.~\ref{fig:gallery_vs_centroid}.
Additionally, we projected the prototypes onto the same embedding space,
highlighting their significant distances from the class centroids.

\vspace{-1mm}
\section{Failure Cases}
Fig.~\ref{fig:failed_instances} presents examples where GCP fails to retrieve
the correct gallery instance corresponding to the given query. Despite these
failures, noticeable visual similarities can be observed between the query
images and the retrieved results. In some instances (e.g., the middle column of
Fig.~\ref{fig:failed_instances_cuhk} and
Fig.~\ref{fig:failed_instances_market}), distinguishing between the query and
retrieved persons is challenging, even for a human.

\def \imgh{1.13cm}
\def \imgw{0.63cm}
\def \imgsep{0pt}
\begin{figure}
\centering
\begin{subfigure}{0.32\textwidth}
  \centering
  \begin{tikzpicture}
  % cuhk query
  \begin{scope}[shift={(0,0)}]
    \node[inner sep=\imgsep] (cuhkq1) at (0,0){\includegraphics[width=\imgw, height=\imgh]
      {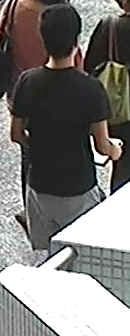}};
    \node[inner sep=\imgsep, right=.1cm of cuhkq1] (cuhkq2) {\includegraphics[width=\imgw, height=\imgh]
      {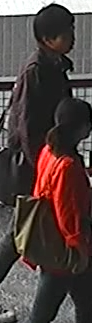}};
    \node[inner sep=\imgsep, right=.1cm of cuhkq2] (cuhkq3) {\includegraphics[width=\imgw, height=\imgh]
      {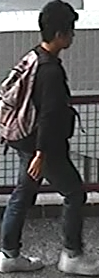}};
    \node[inner sep=\imgsep, right=.1cm of cuhkq3] (cuhkq4) {\includegraphics[width=\imgw, height=\imgh]
      {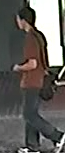}};
    \node[inner sep=\imgsep, right=.1cm of cuhkq4] (cuhkq5){\includegraphics[width=\imgw, height=\imgh]
    {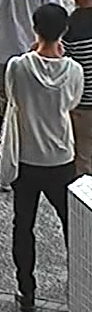}};
    % cuhk top1
    \node[inner sep=\imgsep,below=.1cm of cuhkq1] (cuhkg1) {\includegraphics[width=\imgw, height=\imgh]
      {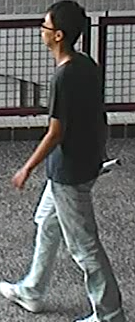}};
    \node[inner sep=\imgsep, right=.1cm of cuhkg1] (cuhkg2) {\includegraphics[width=\imgw, height=\imgh]
      {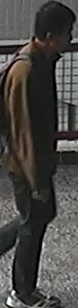}};
    \node[inner sep=\imgsep, right=.1cm of cuhkg2] (cuhkg3) {\includegraphics[width=\imgw, height=\imgh]
      {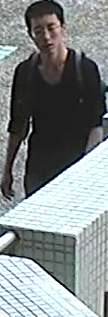}};
    \node[inner sep=\imgsep, right=.1cm of cuhkg3] (cuhkg4) {\includegraphics[width=\imgw, height=\imgh]
      {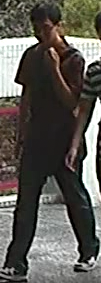}};
    \node[inner sep=\imgsep, right=.1cm of cuhkg4] (cuhkg5){\includegraphics[width=\imgw, height=\imgh]
      {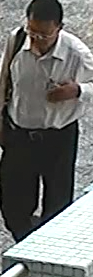}};

    \node[fit=(cuhkq1)(cuhkg5), draw] {};
  \end{scope}
\end{tikzpicture}
\caption{}
\label{fig:failed_instances_cuhk}
\end{subfigure}
\begin{subfigure}{0.32\textwidth}
  \centering
  \begin{tikzpicture}
  % Market query
  \node[inner sep=\imgsep] (marketq1) at (0,0){\includegraphics[width=\imgw, height=\imgh]
    {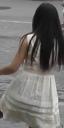}};
  \node[inner sep=\imgsep, right=.1cm of marketq1] (marketq2) {\includegraphics[width=\imgw, height=\imgh]
    {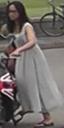}};
  \node[inner sep=\imgsep, right=.1cm of marketq2] (marketq3) {\includegraphics[width=\imgw, height=\imgh]
    {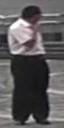}};
  \node[inner sep=\imgsep, right=.1cm of marketq3] (marketq4) {\includegraphics[width=\imgw, height=\imgh]
    {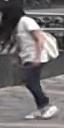}};
  \node[inner sep=\imgsep, right=.1cm of marketq4] (marketq5){\includegraphics[width=\imgw, height=\imgh]
    {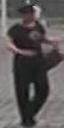}};

  % Gallery
  \node[inner sep=\imgsep,below=.1cm of marketq1] (marketg1) {\includegraphics[width=\imgw, height=\imgh]
    {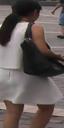}};
  \node[inner sep=\imgsep, right=.1cm of marketg1] (marketg2) {\includegraphics[width=\imgw, height=\imgh]
    {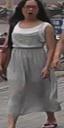}};
  \node[inner sep=\imgsep, right=.1cm of marketg2] (marketg3) {\includegraphics[width=\imgw, height=\imgh]
    {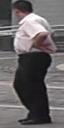}};
  \node[inner sep=\imgsep, right=.1cm of marketg3] (marketg4) {\includegraphics[width=\imgw, height=\imgh]
    {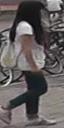}};
  \node[inner sep=\imgsep, right=.1cm of marketg4] (marketg5){\includegraphics[width=\imgw, height=\imgh]
    {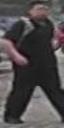}};
  \node[fit=(marketq1)(marketg5), draw] {};
\end{tikzpicture}
\caption{}
\label{fig:failed_instances_market}
\end{subfigure}
\begin{subfigure}{0.32\textwidth}
  \centering
  \begin{tikzpicture}
  % Market query
  \node[inner sep=\imgsep] (msmtq1) at (0,0){\includegraphics[width=\imgw, height=\imgh]
    {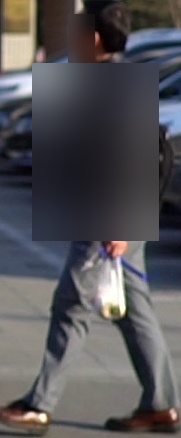}};
  \node[inner sep=\imgsep, right=.1cm of msmtq1] (msmtq2) {\includegraphics[width=\imgw, height=\imgh]
    {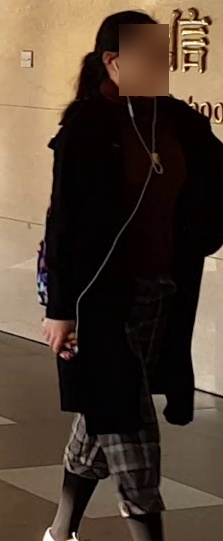}};
  \node[inner sep=\imgsep, right=.1cm of msmtq2] (msmtq3) {\includegraphics[width=\imgw, height=\imgh]
    {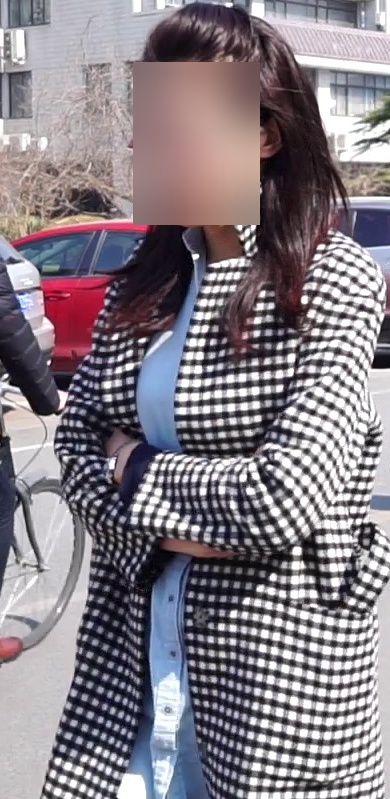}};
  \node[inner sep=\imgsep, right=.1cm of msmtq3] (msmtq4) {\includegraphics[width=\imgw, height=\imgh]
    {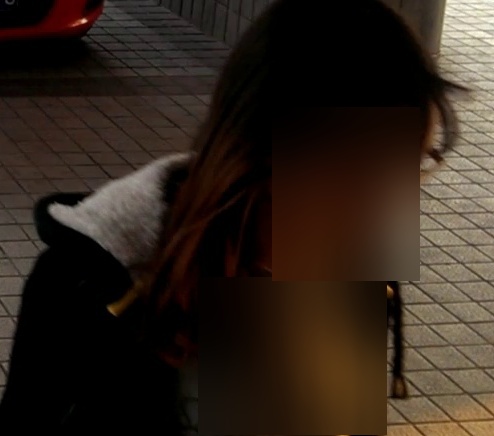}};
  \node[inner sep=\imgsep, right=.1cm of msmtq4] (msmtq5){\includegraphics[width=\imgw, height=\imgh]
    {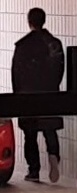}};

  % Gallery
  \node[inner sep=\imgsep, below=.1cm of msmtq1] (msmtg1) {\includegraphics[width=\imgw, height=\imgh]
    {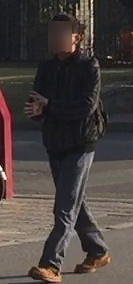}};
  \node[inner sep=\imgsep, right=.1cm of msmtg1] (msmtg2) {\includegraphics[width=\imgw, height=\imgh]
    {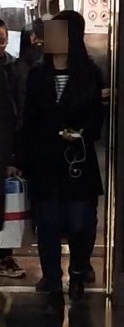}};
  \node[inner sep=\imgsep, right=.1cm of msmtg2] (msmtg3) {\includegraphics[width=\imgw, height=\imgh]
    {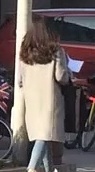}};
  \node[inner sep=\imgsep, right=.1cm of msmtg3] (msmtg4) {\includegraphics[width=\imgw, height=\imgh]
    {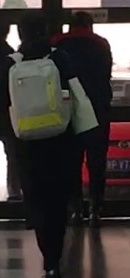}};
  \node[inner sep=\imgsep, right=.1cm of msmtg4] (msmtg5){\includegraphics[width=\imgw, height=\imgh]
    {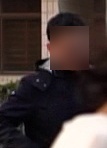}};
 \node[fit=(msmtq1)(msmtg5), draw] {};
\end{tikzpicture}
\caption{}
\label{fig:failed_instances_msmt}
\end{subfigure}
\caption{Examples where GCP fails at R-1 retrieval on the (a) CUHK02-NP, (b)
Market1501, and (c) MSMT17 datasets. The top row displays the query images,
while the images directly below represent the R-1 retrievals from the gallery.}
\label{fig:failed_instances}
\vspace{-4mm}
\end{figure}
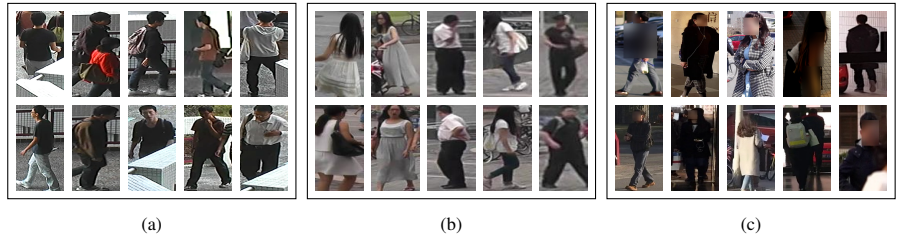

\end{document}